\documentclass[11pt,a4paper]{article}
\usepackage[nohyperref]{naaclhlt2018}
\usepackage{times}
\usepackage{latexsym}
\usepackage{url}
\usepackage{booktabs}       

\usepackage[utf8]{inputenc}
\usepackage[T1]{fontenc}
\usepackage{amsmath}
\usepackage{multirow}
\usepackage{graphicx}
\usepackage{bm}
\usepackage{amsmath,amsfonts,amssymb}
\usepackage{subcaption}

\usepackage{xcolor}
\definecolor{nice-red}{HTML}{E41A1C}
\definecolor{nice-orange}{HTML}{FF7F00}
\definecolor{nice-yellow}{HTML}{FFC020}
\definecolor{nice-green}{HTML}{4DAF4A}
\definecolor{nice-blue}{HTML}{377EB8}
\definecolor{nice-purple}{HTML}{984EA3}

\usepackage{array}
\newcolumntype{L}{>{\centering\arraybackslash}m{1\linewidth}}

\aclfinalcopy 

\newcommand*{\affaddr}[1]{#1}
\newcommand*{\affmark}[1][*]{\textsuperscript{#1}}
\newcommand*{\email}[1]{\texttt{#1}}

\title{Multi-task Learning of Pairwise Sequence Classification Tasks\\Over Disparate Label Spaces}

\author{Isabelle Augenstein\affmark[1]\footnotemark, Sebastian Ruder\affmark[2]\affmark[3]\footnotemark[1], Anders S{\o}gaard\affmark[1]\\
\affaddr{\affmark[1]Department of Computer Science, University of Copenhagen, Denmark}\\
\affaddr{\affmark[2]Insight Research Centre, National University of Ireland, Galway}\\
\affaddr{\affmark[3]Aylien Ltd., Dublin, Ireland}\\
\email{\{augenstein|soegaard\}@di.ku.dk}, \email{sebastian@ruder.io}  
}

\date{}

\begin{document}
\maketitle
\begin{abstract}
We combine multi-task learning and semi-supervised learning by inducing a joint embedding space between disparate label spaces and learning transfer functions between label embeddings, enabling us to jointly leverage unlabelled data and auxiliary, annotated datasets. We evaluate our approach on a variety of sequence classification tasks with disparate label spaces. We outperform strong single and multi-task baselines and achieve a new state-of-the-art for topic-based sentiment analysis.
\end{abstract}

\newenvironment{starfootnotes}
  {\par\edef\savedfootnotenumber{\number\value{footnote}}
   \renewcommand{\thefootnote}{$\star$} 
   \setcounter{footnote}{0}}
  {\par\setcounter{footnote}{\savedfootnotenumber}}

\begin{starfootnotes}
\footnotetext{The first two authors contributed equally.}
\end{starfootnotes}

\section{Introduction}

Multi-task learning (MTL) and semi-supervised learning are both successful paradigms for learning in scenarios with limited labelled data and have in recent years been applied to almost all areas of NLP. Applications of MTL in NLP, for example, include partial parsing \cite{Soegaard:Goldberg:16}, text normalisation \cite{Bollman:ea:17}, neural machine translation \cite{Luong:ea:16}, and keyphrase boundary classification \citep{Augenstein2017KBC}.

Contemporary work in MTL for NLP typically focuses on learning representations that are useful across tasks, often through hard parameter sharing of hidden layers of neural networks \cite{Collobert2011,Soegaard:Goldberg:16}. If tasks share optimal hypothesis classes at the level of these representations, MTL leads to improvements \cite{Baxter:00}. However, while sharing hidden layers of neural networks is an effective regulariser \cite{Soegaard:Goldberg:16}, we potentially {\em loose synergies between the classification functions} trained to associate these representations with class labels. This paper sets out to build an architecture in which such synergies are exploited, with an application to pairwise sequence classification tasks. Doing so, we achieve a new state of the art on topic-based sentiment analysis.

For many NLP tasks, disparate label sets are weakly correlated, e.g. part-of-speech tags correlate with dependencies \cite{Hashimoto2017}, sentiment correlates with emotion \cite{Felbo2017,EisnerEmoji}, etc. We thus propose to induce a joint label embedding space (visualised in Figure \ref{fig:label_embeddings}) using a Label Embedding Layer that allows us to model these relationships, which we show helps with learning.

In addition, for tasks where labels are closely related, we should be able to not only model their relationship, but also to directly estimate the corresponding label of the target task based on auxiliary predictions. To this end, we propose to train a Label Transfer Network (LTN) jointly with the model to produce pseudo-labels across tasks. 

The LTN can be used to label unlabelled and auxiliary task data by utilising the `dark knowledge' \cite{Hinton2015} contained in auxiliary model predictions. This pseudo-labelled data is then incorporated into the model via semi-supervised learning, leading to a natural combination of multi-task learning and semi-supervised learning. We additionally augment the LTN with data-specific diversity features \cite{ruder2017emnlp} that aid in learning. 

\paragraph{Contributions} Our contributions are: a) We model the relationships between labels by inducing a joint label space for multi-task learning. b) We propose a Label Transfer Network that learns to transfer labels between tasks and propose to use semi-supervised learning to leverage them for training. c) We evaluate MTL approaches on a variety of classification tasks and shed new light on settings where multi-task learning works. d) We perform an extensive ablation study of our model. e) We report state-of-the-art performance on topic-based sentiment analysis.

\section{Related work}

\paragraph{Learning task similarities} Existing approaches for learning similarities between tasks enforce a clustering of tasks \cite{Evgeniou2005,Jacob2009}, induce a shared prior \cite{Yu2005,Xue2007,DaumeIII2009}, or learn a grouping \cite{Kang2011,Kumar2012}. These approaches focus on homogeneous tasks and employ linear or Bayesian models. They can thus not be directly applied to our setting with tasks using disparate label sets.

\paragraph{Multi-task learning with neural networks} Recent work in multi-task learning goes beyond hard parameter sharing~\citep{Caruana:93} and considers different sharing structures, e.g. only sharing at lower layers \citep{Soegaard:Goldberg:16} and induces private and shared subspaces \citep{Liu2017,ruder2017sluice}. These approaches, however, are not able to take into account relationships between labels that may aid in learning. Another related direction is to train on disparate annotations of the same task \cite{chen-zhang-liu:2016:EMNLP2016,Peng2017}. In contrast, the different nature of our tasks requires a modelling of their label spaces.

\paragraph{Semi-supervised learning} There exists a wide range of semi-supervised learning algorithms, e.g., self-training, co-training, tri-training, EM, and combinations thereof, several of which have also been used in NLP. Our approach is probably most closely related to an algorithm called {\em co-forest} \cite{Li:Zhou:07}. In co-forest, like here, each learner is improved with unlabeled instances labeled by the ensemble consisting of all the other learners. 
Note also that several researchers have proposed using auxiliary tasks that are unsupervised \cite{Plank2016a,Rei2017}, which also leads to a form of semi-supervised models. 

\paragraph{Label transformations} The idea of manually mapping between label sets or learning such a mapping to facilitate transfer is not new. \newcite{Zhang:ea:12} use distributional information to map from a language-specific tagset to a tagset used for other languages, in order to facilitate cross-lingual transfer. More related to this work, \newcite{Kim:ea:15} use canonical correlation analysis to transfer between tasks with disparate label spaces. There has also been work on label transformations in the context of multi-label classification problems \cite{Yeh:ea:17}.


\section{Multi-task learning with disparate label spaces}

\subsection{Problem definition}

In our multi-task learning scenario, we have access to labelled datasets for $T$ tasks $\mathcal{T}_1, \ldots, \mathcal{T}_T$ at training time with a target task $\mathcal{T}_T$ that we particularly care about. The training dataset for task $\mathcal{T}_i$ consists of $N_k$ examples $X_{\mathcal{T}_i} = \{x_1^{\mathcal{T}_i}, \ldots, x_{N_k}^{\mathcal{T}_i}\}$ and their labels $Y_{\mathcal{T}_i} = \{\mathbf{y}_1^{\mathcal{T}_i}, \ldots, \mathbf{y}_{N_k}^{\mathcal{T}_i}\}$.
Our base model is a deep neural network that performs classic hard parameter sharing \cite{Caruana:93}: It shares its parameters across tasks and has task-specific softmax output layers, which output a probability distribution $\mathbf{p}^{\mathcal{T}_i}$ for task $\mathcal{T}_i$ according to the following equation:

\begin{equation}
\mathbf{p}^{\mathcal{T}_i} = \mathrm{softmax}(\mathbf{W}^{\mathcal{T}_i}\mathbf{h} + \mathbf{b}^{\mathcal{T}_i})
\end{equation}

where $\mathrm{softmax}(\mathbf{x}) = e^\mathbf{x} / \sum^{ \|\mathbf{x}\| }_{i=1} e^{\mathbf{x}_i}$, $\mathbf{W}^{\mathcal{T}_i} \in \mathbb{R}^{L_i \times h}$, $\mathbf{b}^{\mathcal{T}_i} \in \mathbb{R}^{L_i}$ is the weight matrix and bias term of the output layer of task $\mathcal{T}_i$ respectively, $\mathbf{h} \in \mathbb{R}^h$ is the jointly learned hidden representation, $L_i$ is the number of labels for task $\mathcal{T}_i$, and $h$ is the dimensionality of $\mathbf{h}$.

\begin{figure*}[!htb]
    \begin{subfigure}{.32\linewidth}
      \centering
         \includegraphics[height=2.2in]{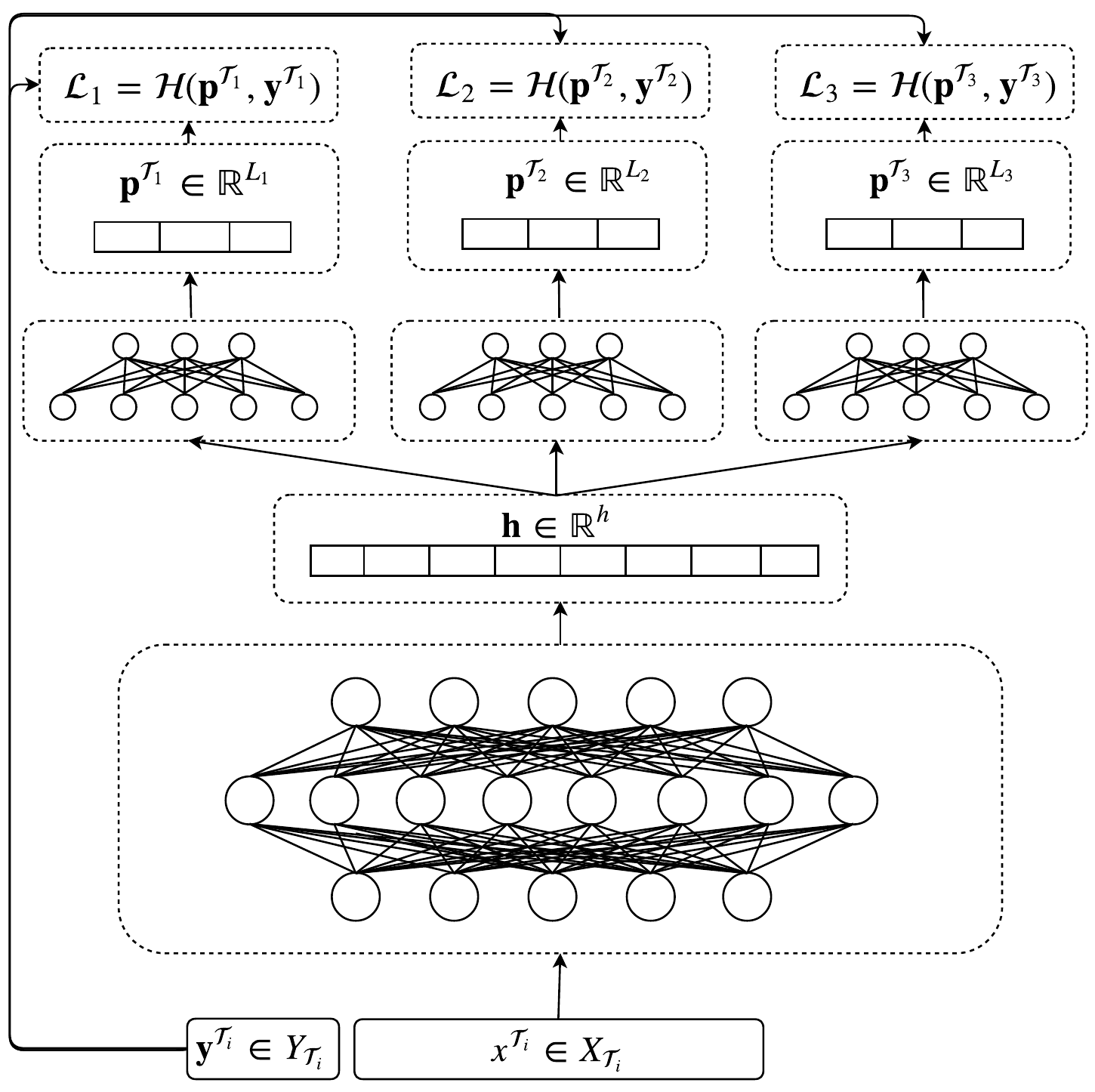}
    \caption{Multi-task learning} \label{fig:mtl}
    \end{subfigure}%
    \hspace*{0.4cm}
    \begin{subfigure}{.21\linewidth}
      \centering
         \includegraphics[height=2.2in]{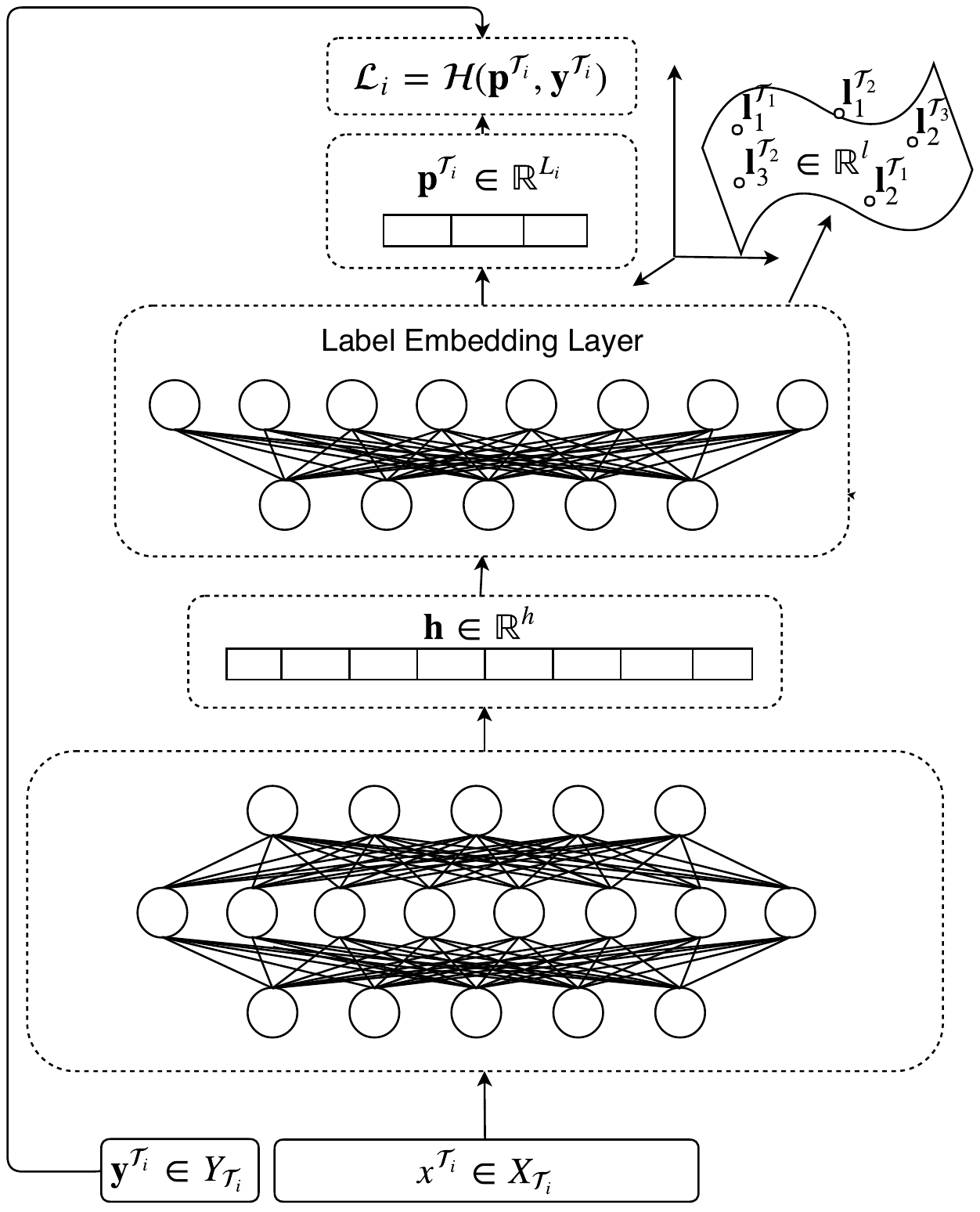}
    \caption{MTL with LEL} \label{fig:lel}
    \end{subfigure}
    \hspace*{0.4cm}
    \begin{subfigure}{.45\linewidth}
      \centering
         \includegraphics[height=2.2in]{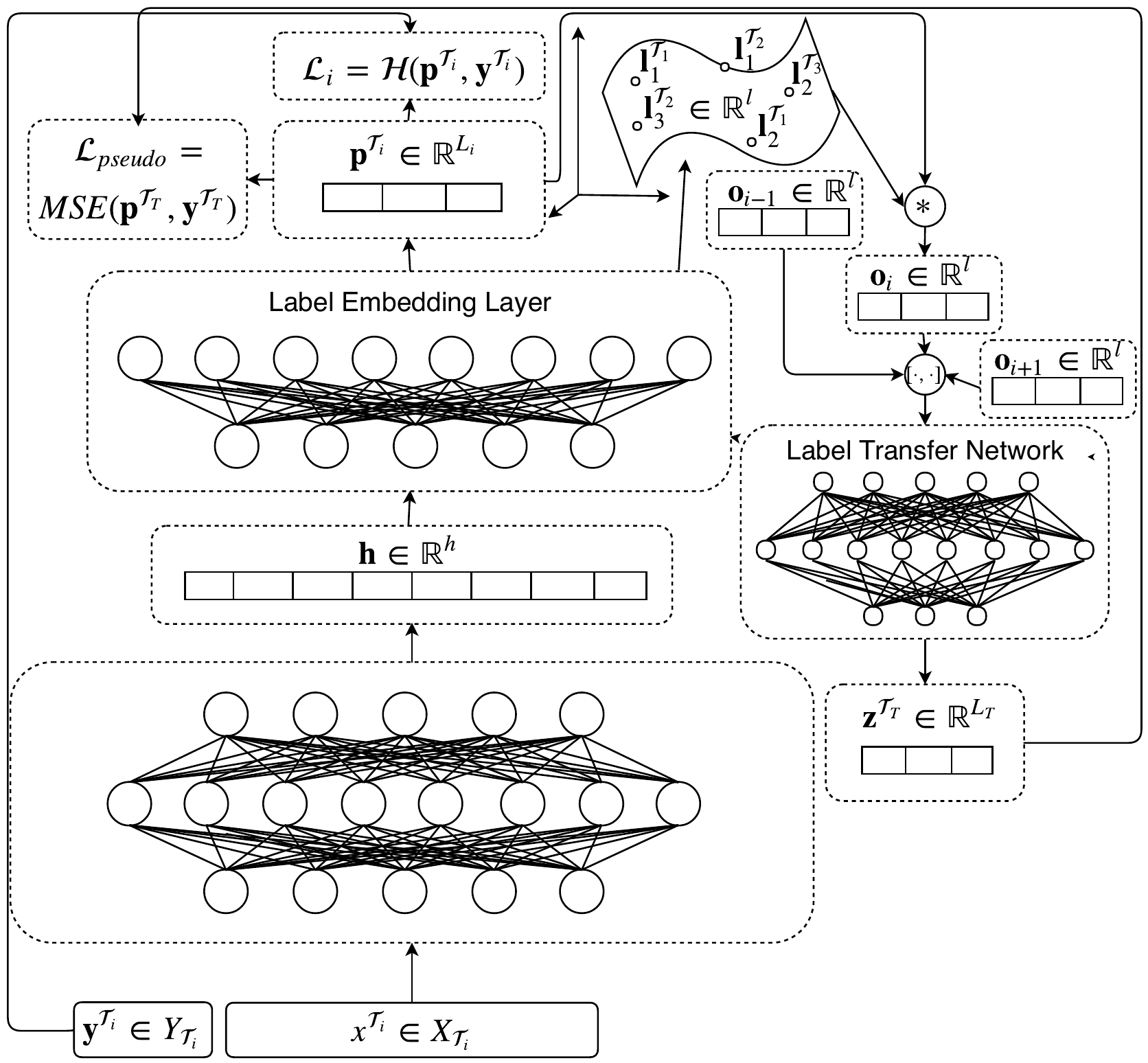}
    \caption{Semi-supervised MTL with LTN} \label{fig:semi-supervised_mtl}
    \end{subfigure}
    \caption{a) Multi-task learning (MTL) with hard parameter sharing and 3 tasks $\mathcal{T}_{1-3}$ and $L_{1-3}$ labels per task. A shared representation $\mathbf{h}$ is used as input to task-specific softmax layers, which optimise cross-entropy losses $\mathcal{L}_{1-3}$. b) MTL with the Label Embedding Layer (LEL) embeds task labels $\mathbf{l}_{1-L_i}^{\mathcal{T}_{1-3}}$ in a joint embedding space and uses these for prediction with a label compatibility function. c) Semi-supervised MTL with the Label Transfer Network (LTN) in addition optimises an unsupervised loss $\mathcal{L}_{pseudo}$ over pseudo-labels $\mathbf{z}^{\mathcal{T}_T}$ on auxiliary/unlabelled data. The pseudo-labels $\mathbf{z}^{\mathcal{T}_T}$ are produced by the LTN for the main task $\mathcal{T}_T$ using the concatenation of auxiliary task label output embeddings $[\mathbf{o}_{i-1},\mathbf{o}_i, \mathbf{o}_{i+1}]$ as input.}
\label{fig:training-procedures}
\end{figure*}

The MTL model is then trained to minimise the sum of the individual task losses:

\begin{equation} \label{eq:mtl_loss}
\mathcal{L} = \lambda_1 \mathcal{L}_1 + \ldots + \lambda_T \mathcal{L}_T
\end{equation}

where $\mathcal{L}_i$ is the negative log-likelihood objective $\mathcal{L}_i = \mathcal{H}(\mathbf{p}^{\mathcal{T}_i},\mathbf{y}^{\mathcal{T}_i}) = - \frac{1}{N} \sum_n \sum_j \log \mathbf{p}_j^{\mathcal{T}_i} \mathbf{y}_j^{\mathcal{T}_i} $ and $\lambda_i$ is a parameter that determines the weight of task $\mathcal{T}_i$. In practice, we apply the same weight to all tasks. We show the full set-up in Figure \ref{fig:mtl}.

\subsection{Label Embedding Layer}

In order to learn the relationships between labels, we propose a Label Embedding Layer (LEL) that embeds the labels of all tasks in a joint space. Instead of training separate softmax output layers as above, we introduce a label compatibility function $c(\cdot, \cdot)$ that measures how similar a label with embedding $\mathbf{l}$ is to the hidden representation $\mathbf{h}$:

\begin{equation}
c(\mathbf{l},\mathbf{h}) = \mathbf{l} \cdot \mathbf{h}
\end{equation}

where $\cdot$ is the dot product. This is similar to the Universal Schema Latent Feature Model introduced by \newcite{Riedel2013}. In contrast to other models that use the dot product in the objective function, we do not have to rely on negative sampling and a hinge loss \cite{Collobert2008} as negative instances (labels) are known. For efficiency purposes, we use matrix multiplication instead of a single dot product and softmax instead of sigmoid activations:

\begin{equation}
\mathbf{p} = \mathrm{softmax}(\mathbf{L} \mathbf{h})
\end{equation}

where $\mathbf{L} \in \mathbb{R}^{(\sum_i L_i) \times l}$ is the label embedding matrix for all tasks and $l$ is the dimensionality of the label embeddings. In practice, we set $l$ to the hidden dimensionality $h$. We use padding if $l < h$. We apply a task-specific mask to $\mathbf{L}$ in order to obtain a task-specific probability distribution $\mathbf{p}^{\mathcal{T}_i}$. The LEL is shared across all tasks, which allows us to learn the relationships between the labels in the joint embedding space. We show MTL with the LEL in Figure \ref{fig:lel}.

\subsection{Label Transfer Network}

The LEL allows us to learn the relationships between labels. In order to make use of these relationships, we would like to leverage the predictions of our auxiliary tasks to estimate a label for the target task. To this end, we introduce the Label Transfer Network (LTN). This network takes the auxiliary task outputs as input. In particular, we define the output label embedding $\mathbf{o}_i$ of task $\mathcal{T}_i$ as the sum of the task's label embeddings $\mathbf{l}_j$ weighted with their probability $\mathbf{p}^{\mathcal{T}_i}_j$:

\begin{equation}
\mathbf{o}_i = \sum^{L_i}_{j=1} \mathbf{p}^{\mathcal{T}_i}_j \mathbf{l}_j 
\end{equation}

The label embeddings $\mathbf{l}$ encode general relationship between labels, while the model's probability distribution $\mathbf{p}^{\mathcal{T}_i}$ over its predictions encodes fine-grained information useful for learning \cite{Hinton2015}. The LTN is trained on labelled target task data. For each example, the corresponding label output embeddings of the auxiliary tasks are fed into a  multi-layer perceptron (MLP), which is trained with a negative log-likelihood objective $\mathcal{L}_\mathrm{LTN}$ to produce a pseudo-label $\mathbf{z}^{\mathcal{T}_T}$ for the target task $\mathcal{T}_{T}$:

\begin{equation}
\mathrm{LTN}_T = \mathrm{MLP}([\mathbf{o}_1, \ldots, \mathbf{o}_{T-1}]) 
\end{equation}

where $[\cdot, \cdot]$ designates concatenation. The mapping of the tasks in the LTN yields another signal that can be useful for optimisation and act as a regulariser. The LTN can also be seen as a mixture-of-experts layer \cite{Jacobs1991} where the experts are the auxiliary task models. As the label embeddings are learned jointly with the main model, the LTN is more sensitive to the relationships between labels than a separately learned mixture-of-experts model that only relies on the experts' output distributions. As such, the LTN can be directly used to produce predictions on unseen data.

\subsection{Semi-supervised MTL}

The downside of the LTN is that it requires additional parameters and relies on the predictions of the auxiliary models, which impacts the runtime during testing. Instead, of using the LTN for prediction directly, we can use it to provide pseudo-labels for unlabelled or auxiliary task data by utilising auxiliary predictions for semi-supervised learning.

We train the target task model on the pseudo-labelled data to minimise the squared error between the model predictions $\mathbf{p}^{\mathcal{T}_i}$ and the pseudo labels $\mathbf{z}^{\mathcal{T}_i}$ produced by the LTN:

\begin{equation}
\mathcal{L}_{pseudo} = MSE(\mathbf{p}^{\mathcal{T}_T}, \mathbf{z}^{\mathcal{T}_T}) = ||\mathbf{p}^{\mathcal{T}_T} - \mathbf{z}^{\mathcal{T}_T}|| ^ 2
\end{equation}

We add this loss term to the MTL loss in Equation \ref{eq:mtl_loss}.
As the LTN is learned together with the MTL model, pseudo-labels produced early during training will likely not be helpful as they are based on unreliable auxiliary predictions. For this reason, we first train the base MTL model until convergence and then augment it with the LTN.
We show the full semi-supervised learning procedure in Figure \ref{fig:semi-supervised_mtl}.

\subsection{Data-specific features}

When there is a domain shift between the datasets of different tasks as is common for instance when learning NER models with different label sets, the output label embeddings might not contain sufficient information to bridge the domain gap.

To mitigate this discrepancy, we augment the LTN's input with features that have been found useful for transfer learning \cite{ruder2017emnlp}. In particular, we use 
the number of word types, type-token ratio, entropy, Simpson's index, and Rényi entropy as diversity features. We calculate each feature for each example.
\footnote{For more information regarding the feature calculation, refer to \newcite{ruder2017emnlp}.} The features are then concatenated with the input of the LTN. 

\subsection{Other multi-task improvements}

Hard parameter sharing can be overly restrictive and provide a regularisation that is too heavy when jointly learning many tasks. For this reason, we propose several additional improvements that seek to alleviate this burden: We use skip-connections, which have been shown to be useful for multi-task learning in recent work \cite{ruder2017sluice}. Furthermore, we add a task-specific layer before the output layer, which is useful for learning task-specific transformations of the shared representations \cite{Soegaard:Goldberg:16,ruder2017sluice}.

\section{Experiments}

For our experiments, we evaluate on a wide range of text classification tasks. In particular, we choose pairwise classification tasks---i.e. those that condition the reading of one sequence on another sequence---as we are interested in understanding if knowledge can be transferred even for these more complex interactions. To the best of our knowledge, this is the first work on transfer learning between such pairwise sequence classification tasks. We implement all our models in Tensorflow \cite{abadi2016tensorflow} and release the code at \url{https://github.com/coastalcph/mtl-disparate}. 

\begin{table}[t] 
\centering
\begin{tabular}{l l c c c}
\toprule
Task & Domain & $N$ & $L$ & Metric\\
\midrule
{\tt Topic-2} & Twitter & 4,346 & 2 & $\rho^{PN}$ \\
{\tt Topic-5} & Twitter & 6,000 & 5 & $MAE^M$\\
{\tt Target} & Twitter & 6,248 & 3 & $F_1^M$ \\
{\tt Stance} & Twitter & 2,914 & 3 & $F_1^{FA}$\\
{\tt ABSA-L} & Reviews & 2,909 & 3 & $Acc$\\
{\tt ABSA-R} & Reviews & 2,507 & 3 & $Acc$\\
{\tt FNC-1} & News & 39,741 & 4 & $Acc$\\
{\tt MultiNLI} & Diverse & 392,702 & 3 & $Acc$\\
\bottomrule
\end{tabular}%
\caption{Training set statistics and evaluation metrics of every task. $N$: \# of examples. $L$: \# of labels.}
\label{tab:dataset-stats}
\end{table}

\subsection{Tasks and datasets}\label{sec:datasets}

\setlength{\tabcolsep}{0.3em}
\begin{table}[h]
\fontsize{10}{10}\selectfont
\begin{center}
\begin{tabular}{|L|}
\toprule
\textbf{Topic-based sentiment analysis}: \\
\textit{Tweet}: No power at home, sat in the dark listening to AC/DC in the hope it'll make the electricity come back again \\
\textit{Topic}: AC/DC \\
\textit{Label}: positive \\
\midrule
\textbf{Target-dependent sentiment analysis}: \\
\textit{Text}: how do you like settlers of catan for the wii? \\
\textit{Target}: wii  \\
\textit{Label}: neutral \\
\midrule
\textbf{Aspect-based sentiment analysis}: \\
\textit{Text}: For the price, you cannot eat this well in Manhattan\\
\textit{Aspects}: restaurant prices, food quality \\
\textit{Label}: positive\\
\midrule
\textbf{Stance detection}: \\
\textit{Tweet}: Be prepared - if we continue the policies of the liberal left, we will be \#Greece\\
\textit{Target}: Donald Trump\\
\textit{Label}: favor\\
\midrule
\textbf{Fake news detection}: \\
\textit{Document}: Dino Ferrari hooked the whopper wels catfish, (...), which could be the biggest in the world.\\
\textit{Headline}: Fisherman lands 19 STONE catfish which could be the biggest in the world to be hooked\\
\textit{Label}: agree\\
\midrule
\textbf{Natural language inference}: \\
\textit{Premise}: Fun for only children\\
\textit{Hypothesis}: Fun for adults and children\\
\textit{Label}: contradiction\\
\bottomrule
\end{tabular}
\end{center}
\caption{\label{tab:dataset-examples} Example instances from the datasets described in Section \ref{sec:datasets}.}
\end{table}

We use the following tasks and datasets for our experiments, show task statistics in Table \ref{tab:dataset-stats}, and summarise examples in Table \ref{tab:dataset-examples}:

\paragraph{Topic-based sentiment analysis} Topic-based sentiment analysis aims to estimate the sentiment of a tweet known to be about a given topic. We use the data from SemEval-2016 Task 4 Subtask B and C \cite{SemEval:2016:task4} for predicting on a two-point scale of positive and negative ({\tt Topic-2}) and five-point scale ranging from highly negative to highly positive ({\tt Topic-5}) respectively. An example from this dataset would be to classify the tweet ``No power at home, sat in the dark listening to AC/DC in the hope it'll make the electricity come back again'' known to be about the topic ``AC/DC'', which is labelled as a positive sentiment. The evaluation metrics for {\tt Topic-2} and {\tt Topic-5} are macro-averaged recall ($\rho^{PN}$) and macro-averaged mean absolute error ($MAE^M$) respectively, which are both averaged across topics.

\paragraph{Target-dependent sentiment analysis} Target-dependent sentiment analysis ({\tt Target}) seeks to classify the sentiment of a text's author towards an entity that occurs in the text as positive, negative, or neutral. We use the data from \citeauthor{Dong2014} \shortcite{Dong2014}. An example instance is the expression ``how do you like settlers of catan for the wii?'' which is labelled as neutral towards the target ``wii'.' The evaluation metric is macro-averaged $F_1$ ($F_1^M$).

\paragraph{Aspect-based sentiment analysis} Aspect-based sentiment analysis is the task of identifying whether an aspect, i.e. a particular property of an item is associated with a positive, negative, or neutral sentiment \cite{Ruder2016a}. We use the data of SemEval-2016 Task 5 Subtask 1 Slot 3 \cite{Pontiki2016Aspect} for the laptops ({\tt ABSA-L}) and restaurants ({\tt ABSA-R}) domains. An example is the sentence ``For the price, you cannot eat this well in Manhattan'', labelled as positive towards both the aspects ``restaurant prices'' and ``food quality''. The evaluation metric for both domains is accuracy ($Acc$).

\paragraph{Stance detection} Stance detection ({\tt Stance}) requires a model, given a text and a target entity, which might not appear in the text, to predict whether the author of the text is in favour or against the target or whether neither inference is likely \cite{Augenstein2016stance}. We use the data of SemEval-2016 Task 6 Subtask B \cite{mohammad-EtAl:2016:SemEval}. An example from this dataset would be to predict the stance of the tweet ``Be prepared - if we continue the policies of the liberal left, we will be \#Greece'' towards the topic ``Donald Trump'', labelled as ``favor''. The evaluation metric is the macro-averaged $F_1$ score of the ``favour'' and ``against'' classes ($F_1^{FA}$).

\begin{table*}[h]
\centering
\resizebox{\textwidth}{!}{%
\begin{tabular}{l c c c c c c c c}
\toprule
 & {\tt Stance} & {\tt FNC} & {\tt MultiNLI} & {\tt Topic-2} & {\tt Topic-5}* & {\tt ABSA-L} & {\tt ABSA-R} & {\tt Target}\\
\midrule
\newcite{Augenstein2016stance} & \textbf{49.01} & - & - & - & - & - & - & -\\
\newcite{Riedel2017} & - & \textbf{88.46} & - & - & - & - & - & -\\
\citeauthor{chen2017recurrent} \shortcite{chen2017recurrent} & - & - & \textbf{74.90} & - & - & - & - & -\\
\citeauthor{palogiannidi2016tweester} \shortcite{palogiannidi2016tweester} & - & - & - & \underline{79.90} & - & - & - & -\\
\citeauthor{balikas2016twise} \shortcite{balikas2016twise} & - & - & - & - & \textbf{0.719} & - & - & - \\
\citeauthor{Brun2016} \shortcite{Brun2016} & - & - & - & - & - & - & \textbf{88.13} & -\\
\citeauthor{Kumar2016} \shortcite{Kumar2016} & - & - & - & - & - & \textbf{82.77} & \underline{86.73} & -\\
\citeauthor{Vo2015} \shortcite{Vo2015} & - & - & - & - & - & - &- & \textbf{69.90} \\
\midrule
STL & 41.1 & 72.72 & 49.25 & 63.92 & 0.919 & \underline{76.74} & 67.47 & 64.01 \\
\midrule
MTL + LEL & \underline{46.26} & 72.71 & \underline{49.94}  & \textbf{80.52} & 0.814 & 74.94 & 79.90 & \underline{66.42} \\
MTL + LEL + LTN, main model & 43.16  & \underline{72.73} &  48.75 & 73.90 & \underline{0.810} & 75.06 & 83.71 & 66.10 \\
MTL + LEL + LTN + semi, main model & 43.56 & 72.72 & 48.00 & 72.35  & 0.821 & 75.42 & 83.26 & 63.00 \\
\bottomrule
\end{tabular}%
}
\caption{Comparison of our best performing models on the test set against a single task baseline and the state of the art, with task specific metrics. *: lower is better. Bold: best. Underlined: second-best.}
\label{tab:results-sota}
\end{table*}

\paragraph{Fake news detection} The goal of fake news detection in the context of the Fake News Challenge\footnote{\url{http://www.fakenewschallenge.org/}} is to estimate whether the body of a news article agrees, disagrees, discusses, or is unrelated towards a headline. We use the data from the first stage of the Fake News Challenge ({\tt FNC-1}). An example for this dataset is the document ``Dino Ferrari hooked the whopper wels catfish, (...), which could be the biggest in the world.'' with the headline ``Fisherman lands 19 STONE catfish which could be the biggest in the world to be hooked'' labelled as ``agree''. The evaluation metric is accuracy ($Acc$)\footnote{We use the same metric as \newcite{Riedel2017}.}.

\paragraph{Natural language inference} Natural language inference is the task of predicting whether one sentences entails, contradicts, or is neutral towards another one. We use the Multi-Genre NLI corpus ({\tt MultiNLI}) from the RepEval 2017 shared task \cite{Nangia2017}. An example for an instance would be the sentence pair ``Fun for only children'', ``Fun for adults and children'', which are in a ``contradiction'' relationship. The evaluation metric is accuracy ($Acc$).

\subsection{Base model}

Our base model is the Bidirectional Encoding model \cite{Augenstein2016stance}, a state-of-the-art model for stance detection that conditions a bidirectional LSTM (BiLSTM) encoding of a text on the BiLSTM encoding of the target. Unlike \newcite{Augenstein2016stance}, we do not pre-train word embeddings on a larger set of unlabelled in-domain text for each task as we are mainly interested in exploring the benefit of multi-task learning for generalisation.

\subsection{Training settings}

We use BiLSTMs with one hidden layer of $100$ dimensions, $100$-dimensional randomly initialised word embeddings, a label embedding size of $100$. We train our models with RMSProp, a learning rate of $0.001$, a batch size of $128$, and early stopping on the validation set of the main task with a patience of $3$.

\section{Results}

Our main results are shown in Table \ref{tab:results-sota}, with a comparison against the state of the art. We present the results of our multi-task learning network with label embeddings (MTL + LEL), multi-task learning with label transfer (MTL + LEL + LTN), and the semi-supervised extension of this model. On 7/8 tasks, at least one of our architectures is better than single-task learning; and in 4/8, all our architectures are much better than single-task learning. 

The state-of-the-art systems we compare against are often highly specialised, task-dependent architectures. Our architectures, in contrast, have not been optimised to compare favourably against the state of the art, as our main objective is to develop a novel approach to multi-task learning leveraging synergies between label sets and knowledge of marginal distributions from unlabeled data. For example, we do not use pre-trained word embeddings \cite{Augenstein2016stance,palogiannidi2016tweester,Vo2015}, class weighting to deal with label imbalance \cite{balikas2016twise}, or domain-specific sentiment lexicons \cite{Brun2016,Kumar2016}. Nevertheless, our approach outperforms the state-of-the-art on two-way topic-based sentiment analysis ({\tt Topic-2}).

The poor performance compared to the state-of-the-art on {\tt FNC} and {\tt MultiNLI} is expected; as we alternate among the tasks during training, our model only sees a comparatively small number of examples of both corpora, which are one and two orders of magnitude larger than the other datasets. For this reason, we do not achieve good performance on these tasks as main tasks, but they are still useful as auxiliary tasks as seen in Table \ref{tab:auxiliary-tasks}.

\section{Analysis}
\subsection{Label Embeddings}

\begin{figure}
	\centering
  	\includegraphics[width=1.0\linewidth]{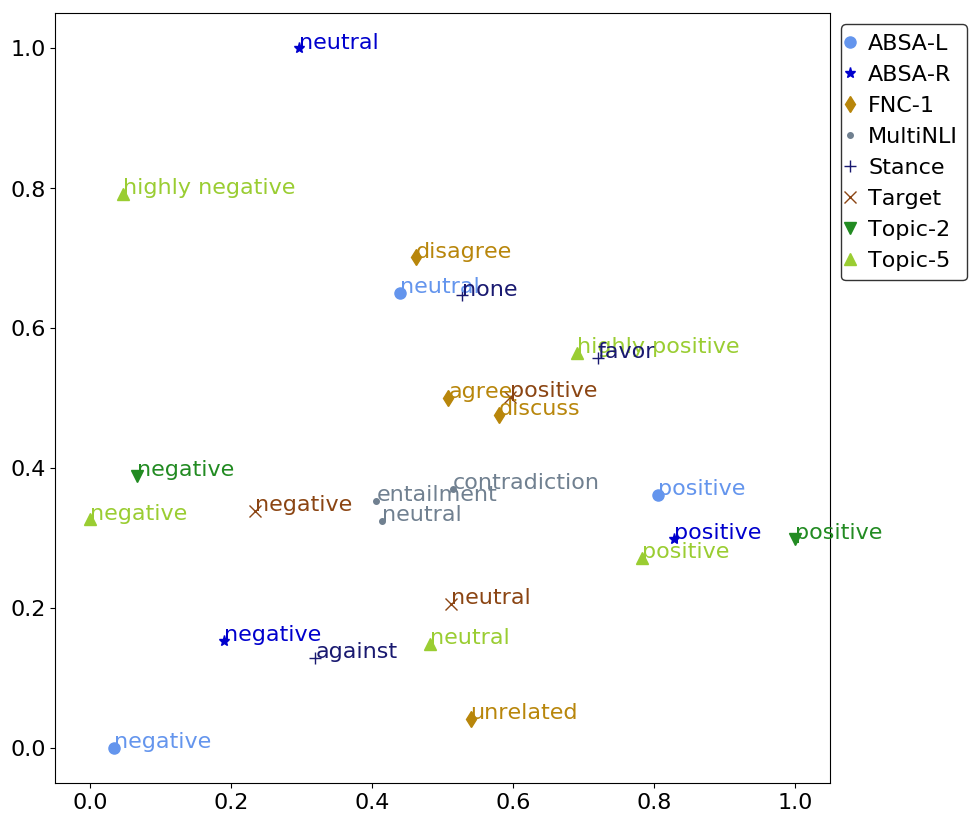}
  	\caption{Label embeddings of all tasks. Positive, negative, and neutral labels are clustered together.}
  	\label{fig:label_embeddings}
\end{figure}

Our results above show that, indeed, modelling the similarity between tasks using label embeddings sometimes leads to much better performance. Figure \ref{fig:label_embeddings} shows why. In Figure~\ref{fig:label_embeddings}, we visualise the label embeddings of an MTL+LEL model trained on all tasks, using PCA. As we can see, similar labels are clustered together across tasks, e.g. there are two positive clusters (middle-right and top-right), two negative clusters (middle-left and bottom-left), and two neutral clusters (middle-top and middle-bottom).


Our visualisation also provides us with a picture of what auxilary tasks are beneficial, and to what extent we can expect synergies from multi-task learning. For instance, the notion of positive sentiment appears to be very similar across the topic-based and aspect-based tasks, while the conceptions of negative and neutral sentiment differ. In addition, we can see that the model has failed to learn a relationship between {\tt MultiNLI} labels and those of other tasks, possibly accounting for its poor performance on the inference task. We did not evaluate the correlation between label embeddings and task performance, but \newcite{Bjerva2017} recently suggested that mutual information of target and auxiliary task label sets is a good predictor of gains from multi-task learning.  

\subsection{Auxilary Tasks}


\begin{table}[h]
\centering
\begin{tabular}{l l}
\toprule
Main task & Auxiliary tasks\\
\midrule
{\tt Topic-2} & {\tt FNC-1}, {\tt MultiNLI}, {\tt Target} \\
\multirow{2}{*}{{\tt Topic-5}} & {\tt FNC-1}, {\tt MultiNLI}, {\tt ABSA-L}, \\
 & {\tt Target}\\
{\tt Target} & {\tt FNC-1}, {\tt MultiNLI}, {\tt Topic-5} \\
{\tt Stance} & {\tt FNC-1}, {\tt MultiNLI}, {\tt Target} \\
{\tt ABSA-L} & {\tt Topic-5} \\
{\tt ABSA-R} & {\tt Topic-5}, {\tt ABSA-L}, {\tt Target}\\
\multirow{2}{*}{{\tt FNC-1}} & {\tt Stance}, {\tt MultiNLI}, {\tt Topic-5},\\
& {\tt ABSA-R}, {\tt Target} \\
{\tt MultiNLI} & {\tt Topic-5} \\
\bottomrule
\end{tabular}%
\caption{Best-performing auxiliary tasks for different main tasks.}
\label{tab:auxiliary-tasks}
\end{table}


For each task, we show the auxiliary tasks that achieved the best performance on the development data in Table \ref{tab:auxiliary-tasks}. In contrast to most existing work, we did not restrict ourselves to performing multi-task learning with only one auxiliary task \cite{Soegaard:Goldberg:16,Bingel:ea:17}. Indeed we find that most often a combination of auxiliary tasks achieves the best performance. In-domain tasks are less used than we assumed; only {\tt Target} is consistently used by all Twitter main tasks. In addition, tasks with a higher number of labels, e.g. {\tt Topic-5} are used more often. Such tasks provide a more fine-grained reward signal, which may help in learning representations that generalise better. Finally, tasks with large amounts of training data such as {\tt FNC-1} and {\tt MultiNLI} are also used more often. Even if not directly related, the larger amount of training data that can be indirectly leveraged via multi-task learning may help the model focus on relevant parts of the representation space \cite{Caruana:93}. These observations shed additional light on when multi-task learning may be useful that go beyond existing studies \cite{Bingel:ea:17}.

\begin{table*}[h]
\centering
\resizebox{\textwidth}{!}{%
\begin{tabular}{l c c c c c c c c}
\toprule
 & {\tt Stance} & {\tt FNC} & {\tt MultiNLI} & {\tt Topic-2} & {\tt Topic-5}* & {\tt ABSA-L} & {\tt ABSA-R} & {\tt Target}\\
\midrule
MTL & 44.12 & \underline{72.75} & \underline{49.39} & \textbf{80.74} & 0.859 & 74.94 & 82.25 & 65.73 \\
\midrule
MTL + LEL & \textbf{46.26} & 72.71 & \textbf{49.94}  & \underline{80.52} & 0.814 & 74.94 & 79.90 & \textbf{66.42} \\
MTL + LTN & 40.95 & 72.72 & 44.14 & 78.31  & 0.851 & 73.98  & 82.37 & 63.71 \\
MTL + LTN, main model & 41.60 & 72.72 & 47.62 & 79.98  & 0.814 & \underline{75.54} & 81.70 & 65.61  \\
MTL + LEL + LTN & \underline{44.48} & \textbf{72.76} & 43.72 & 74.07 & 0.821 & \textbf{75.66} & 81.92 & 65.00 \\
MTL + LEL + LTN, main model & 43.16  & 72.73 &  48.75 & 73.90 & 0.810 & 75.06 & \textbf{83.71} & \underline{66.10} \\
\midrule
MTL + LEL + LTN + main preds feats & 42.78 & 72.72 & 45.41 & 66.30  & 0.835 & 73.86 & 81.81 & 65.08 \\
MTL + LEL + LTN + main preds feats, main model &  42.65 & 72.73 & 48.81 & 67.53  & \textbf{0.803} & 75.18 & 82.59 & 63.95 \\
\midrule
MTL + LEL + LTN + main preds feats -- diversity feats & 42.78 & 72.72 & 43.13 & 66.3 & 0.835 & 73.5 & 81.7 & 63.95 \\
MTL + LEL + LTN + main preds feats -- diversity feats, main model & 42.47 & 72.74 & 47.84 & 67.53 & \underline{0.807} & 74.82 & 82.14 & 65.11 \\
\midrule
MTL + LEL + LTN + semi & 42.65 & \underline{72.75} & 44.28 & 77.81  & 0.841 & 74.10 & 81.36 & 64.45 \\
MTL + LEL + LTN + semi, main model & 43.56 & 72.72 & 48.00 & 72.35  & 0.821 & 75.42 & \underline{83.26} & 63.00 \\

\bottomrule
\end{tabular}%
}
\caption{Ablation results with task-specific evaluation metrics on test set with early stopping on dev set. \textit{LTN} means the output of the relabelling function is shown, which does not use the task predictions, only predictions from other tasks. \textit{LTN + main preds feats} means main model predictions are used as features for the relabelling function. \textit{LTN, main model} means that the main model predictions of the model that trains a relabelling function are used. Note that for {\tt MultiNLI}, we down-sample the training data. *: lower is better. Bold: best. Underlined: second-best.} 
\label{tab:ablation}
\end{table*}

\subsection{Ablation analysis}

We now perform a detailed ablation analysis of our model, the results of which are shown in Table \ref{tab:ablation}. We ablate whether to use the LEL (\textit{+ LEL}), whether to use the LTN (\textit{+ LTN}), whether to use the LEL output or the main model output for prediction (main model output is indicated by \textit{, main model}), and whether to use the LTN as a regulariser or for semi-supervised learning (semi-supervised learning is indicated by \textit{+ semi}). We further test whether to use diversity features (\textit{-- diversity feats}) and whether to use main model predictions for the LTN (\textit{+ main model feats}).

Overall, the addition of the Label Embedding Layer improves the performance over regular MTL in almost all cases.

\begin{table}[h]
\centering
\resizebox{0.5\textwidth}{!}{%
\begin{tabular}{l c c c c }
\toprule
Task & Main & LTN & Main (Semi) & LTN (Semi) \\
\midrule
{\tt Stance} & 2.12 & 2.62 & 1.94 & 1.28 \\
{\tt FNC} & 4.28 & 2.49 & 6.92 & 4.84 \\
{\tt MultiNLI}  & 1.5 & 1.95 & 1.94 & 1.28 \\
{\tt Topic-2} & 6.45 & 4.44 & 5.87 & 5.59 \\
{\tt Topic-5}* & 9.22 & 9.71 & 11.3 & 5.90 \\
{\tt ABSA-L} & 3.79 & 2.52 & 9.06 & 6.63 \\
{\tt ABSA-R} & 10.6 & 6.70 & 9.06 & 6.63 \\
{\tt Target} & 26.3 & 14.6 & 20.1 & 15.7 \\
\bottomrule
\end{tabular}
}
\caption{Error analysis of LTN with and without semi-supervised learning for all tasks. Metric shown: percentage of correct predictions only made by either the relabelling function or the main model, respectively, relative to the the number of all correct predictions.}
\label{tab:ltn-errorana}
\end{table}

\begin{figure}[h]
	\centering
  	\includegraphics[width=1.0\linewidth]{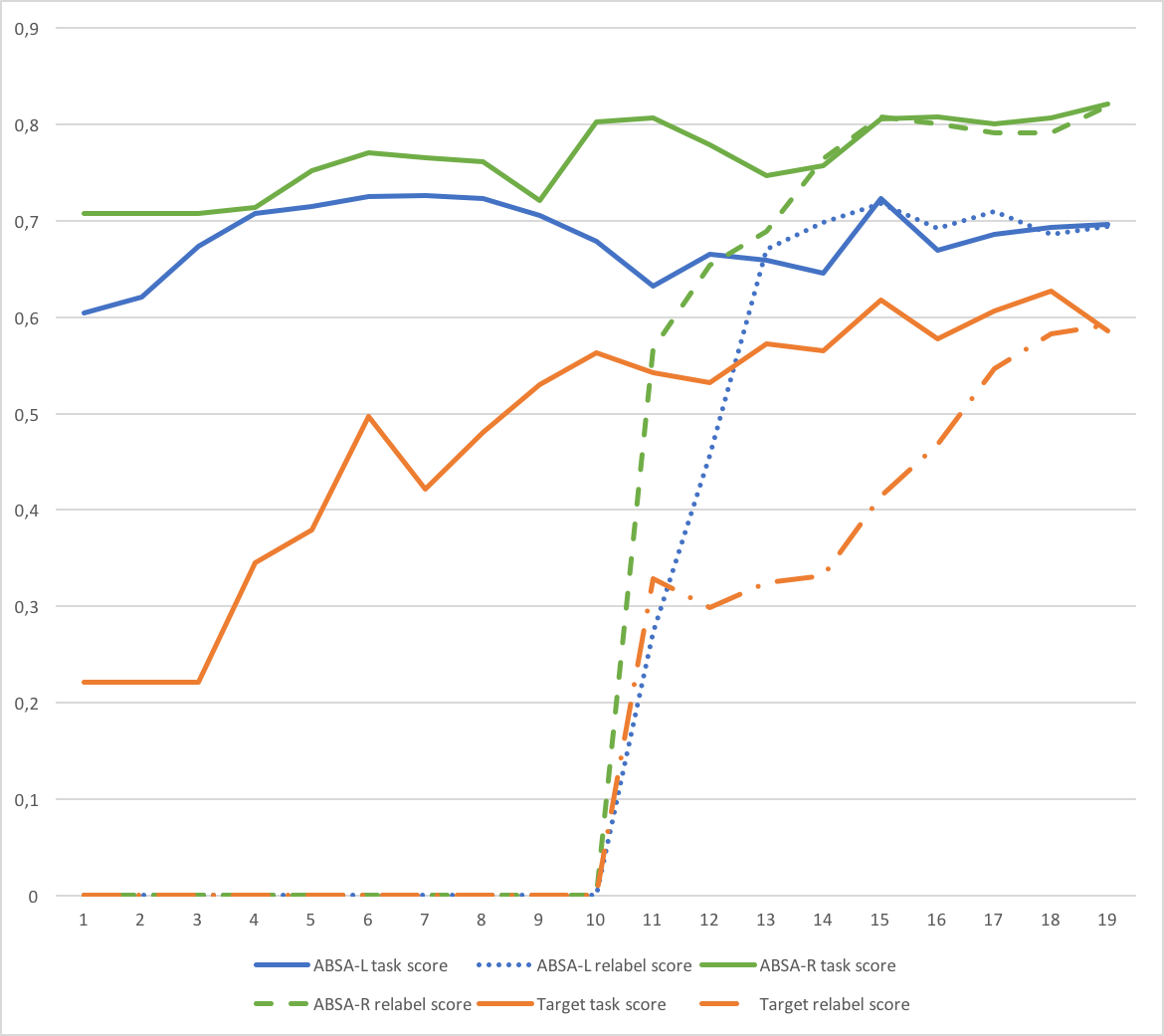}
  	\caption{Learning curves with LTN for selected tasks, dev performances shown. The main model is pre-trained for 10 epochs, after which the relabelling function is trained.}
  	\label{fig:ltn_learning_curve}
\end{figure}

\subsection{Label transfer network}

To understand the performance of the LTN, we analyse learning curves of the relabelling function vs. the main model. Examples for all tasks without semi-supervised learning are shown in Figure \ref{fig:ltn_learning_curve}.
One can observe that the relabelling model does not take long to converge as it has fewer parameters than the main model. Once the relabelling model is learned alongside the main model, the main model performance first stagnates, then starts to increase again. For some of the tasks, the main model ends up with a higher task score than the relabelling model.
We hypothesise that the softmax predictions of other, even highly related tasks are less helpful for predicting main labels than the output layer of the main task model. At best, learning the relabelling model alongside the main model might act as a regulariser to the main model and thus improve the main model's performance over a baseline MTL model, as it is the case for {\tt TOPIC-5} (see Table \ref{tab:ablation}).

To further analyse the performance of the LTN, we look into to what degree predictions of the main model and the relabelling model for individual instances are complementary to one another. Or, said differently, we measure the percentage of correct predictions made only by the relabelling model or made only by the main model, relative to the number of correct predictions overall. Results of this for each task are shown in Table \ref{tab:ltn-errorana} for the LTN with and without semi-supervised learning. One can observe that, even though the relabelling function overall contributes to the score to a lesser degree than the main model, a substantial number of correct predictions are made by the relabelling function that are missed by the main model. This is most prominently pronounced for {\tt ABSA-R}, where the proportion is 14.6.

\section{Conclusion}

We have presented a multi-task learning architecture that (i) leverages potential synergies between classifier functions relating shared representations with disparate label spaces and (ii) enables learning from mixtures of labeled and unlabeled data. We have presented experiments with combinations of eight pairwise sequence classification tasks. Our results show that leveraging synergies between label spaces sometimes leads to big improvements, and we have presented a new state of the art for topic-based sentiment analysis. Our analysis further showed that (a) the learned label embeddings were indicative of gains from multi-task learning, (b) auxiliary tasks were often beneficial across domains,  and (c) label embeddings almost always led to better performance. We also investigated the dynamics of the label transfer network we use for exploiting the synergies between disparate label spaces. 

\section*{Acknowledgments}

Sebastian Ruder is supported by the Irish Research Council Grant Number EBPPG/2014/30 and Science Foundation Ireland Grant Number SFI/12/RC/2289. Anders Søgaard is supported by the ERC Starting Grant Number 313695. Isabelle Augenstein is supported by Eurostars grant Number E10138. We further gratefully acknowledge the support of NVIDIA Corporation with the donation of the Titan Xp GPU used for this research.

\bibliography{semi_multi}

\begin{thebibliography}{}
\expandafter\ifx\csname natexlab\endcsname\relax\def\natexlab#1{#1}\fi

\bibitem[{Abadi et~al.(2016)Abadi, Agarwal, Barham, Brevdo, Chen, Citro,
  Corrado, Davis, Dean, Devin et~al.}]{abadi2016tensorflow}
Mart{\'\i}n Abadi, Ashish Agarwal, Paul Barham, Eugene Brevdo, Zhifeng Chen,
  Craig Citro, Greg~S Corrado, Andy Davis, Jeffrey Dean, Matthieu Devin, et~al.
  2016.
\newblock Tensorflow: Large-scale machine learning on heterogeneous distributed
  systems.
\newblock {\em arXiv preprint arXiv:1603.04467\/} .

\bibitem[{Augenstein et~al.(2016)Augenstein, Rockt{\"a}schel, Vlachos, and
  Bontcheva}]{Augenstein2016stance}
Isabelle Augenstein, Tim Rockt{\"a}schel, Andreas Vlachos, and Kalina
  Bontcheva. 2016.
\newblock {Twitter Stance Detection with Bidirectional Conditional Encoding}.
\newblock In {\em Proceedings of EMNLP\/}.

\bibitem[{Augenstein and S{\o}gaard(2017)}]{Augenstein2017KBC}
Isabelle Augenstein and Anders S{\o}gaard. 2017.
\newblock Multi-task learning of keyphrase boundary detection.
\newblock In {\em Proceedings of ACL\/}.

\bibitem[{Balikas and Amini(2016)}]{balikas2016twise}
Georgios Balikas and Massih-Reza Amini. 2016.
\newblock {TwiSE at SemEval-2016 Task 4: Twitter Sentiment Classification}.
\newblock In {\em Proceedings of SemEval\/}.

\bibitem[{Baxter(2000)}]{Baxter:00}
Jonathan Baxter. 2000.
\newblock {A Model of Inductive Bias Learning}.
\newblock {\em JAIR\/} 12:149--198.

\bibitem[{Bingel and S{\o}gaard(2017)}]{Bingel:ea:17}
Joachim Bingel and Anders S{\o}gaard. 2017.
\newblock Identifying beneficial task relations for multi-task learning in deep
  neural networks.
\newblock In {\em Proceedings of EACL\/}.

\bibitem[{Bjerva(2017)}]{Bjerva2017}
Johannes Bjerva. 2017.
\newblock {Will my auxiliary tagging task help? Estimating Auxiliary Tasks
  Effectivity in Multi-Task Learning}.
\newblock In {\em Proceedings of NODALIDA\/}.

\bibitem[{Bollman et~al.(2017)Bollman, Bingel, and S{\o}gaard}]{Bollman:ea:17}
Marcel Bollman, Joachim Bingel, and Anders S{\o}gaard. 2017.
\newblock Learning attention for historical text normalization by learning to
  pronounce.
\newblock In {\em Proceedings of ACL\/}.

\bibitem[{Brun et~al.(2016)Brun, Perez, and Roux}]{Brun2016}
Caroline Brun, Julien Perez, and Claude Roux. 2016.
\newblock {XRCE at SemEval-2016 Task 5: Feedbacked Ensemble Modelling on
  Syntactico-Semantic Knowledge for Aspect Based Sentiment Analysis}.
\newblock {\em Proceedings of SemEval\/} .

\bibitem[{Caruana(1993)}]{Caruana:93}
Rich Caruana. 1993.
\newblock {Multitask Learning: A Knowledge-Based Source of Inductive Bias}.
\newblock In {\em Proceedings of ICML\/}.

\bibitem[{Chen et~al.(2016)Chen, Zhang, and
  Liu}]{chen-zhang-liu:2016:EMNLP2016}
Hongshen Chen, Yue Zhang, and Qun Liu. 2016.
\newblock {Neural Network for Heterogeneous Annotations}.
\newblock In {\em Proceedings of EMNLP\/}.

\bibitem[{Chen et~al.(2017)Chen, Zhu, Ling, Wei, Jiang, and
  Inkpen}]{chen2017recurrent}
Qian Chen, Xiaodan Zhu, Zhen-Hua Ling, Si~Wei, Hui Jiang, and Diana Inkpen.
  2017.
\newblock Recurrent neural network-based sentence encoder with gated attention
  for natural language inference.
\newblock {\em arXiv preprint arXiv:1708.01353\/} .

\bibitem[{Collobert and Weston(2008)}]{Collobert2008}
Ronan Collobert and Jason Weston. 2008.
\newblock {A Unified Architecture for Natural Language Processing: Deep Neural
  Networks with Multitask Learning}.
\newblock In {\em Proceedings of ICML\/}.

\bibitem[{Collobert et~al.(2011)Collobert, Weston, Bottou, Karlen, Kavukcuoglu,
  and Kuksa}]{Collobert2011}
Ronan Collobert, Jason Weston, L{\'e}on Bottou, Michael Karlen, Koray
  Kavukcuoglu, and Pavel Kuksa. 2011.
\newblock Natural language processing (almost) from scratch.
\newblock {\em The Journal of Machine Learning Research\/} 12:2493--2537.

\bibitem[{{Daum{\'{e}} III}(2009)}]{DaumeIII2009}
Hal {Daum{\'{e}} III}. 2009.
\newblock {Bayesian multitask learning with latent hierarchies}.
\newblock In {\em Proceedings of UAI\/}.

\bibitem[{Dong et~al.(2014)Dong, Wei, Tan, Tang, Zhou, and Xu}]{Dong2014}
Li~Dong, Furu Wei, Chuanqi Tan, Duyu Tang, Ming Zhou, and Ke~Xu. 2014.
\newblock {Adaptive Recursive Neural Network for Target-dependent Twitter
  Sentiment Classification}.
\newblock In {\em Proceedings of ACL\/}. pages 49--54.

\bibitem[{Eisner et~al.(2016)Eisner, Rocktäschel, Augenstein, Bosnjak, and
  Riedel}]{EisnerEmoji}
Ben Eisner, Tim Rocktäschel, Isabelle Augenstein, Matko Bosnjak, and Sebastian
  Riedel. 2016.
\newblock {emoji2vec: Learning Emoji Representations from their Description}.
\newblock In {\em Proceedings of SocialNLP\/}.

\bibitem[{Evgeniou et~al.(2005)Evgeniou, Micchelli, and Pontil}]{Evgeniou2005}
Theodoros Evgeniou, Charles~A. Micchelli, and Massimiliano Pontil. 2005.
\newblock {Learning multiple tasks with kernel methods}.
\newblock {\em Journal of Machine Learning Research\/} 6:615--637.

\bibitem[{Felbo et~al.(2017)Felbo, Mislove, S{\o}gaard, Rahwan, and
  Lehmann}]{Felbo2017}
Bjarke Felbo, Alan Mislove, Anders S{\o}gaard, Iyad Rahwan, and Sune Lehmann.
  2017.
\newblock {Using millions of emoji occurrences to learn any-domain
  representations for detecting sentiment, emotion and sarcasm}.
\newblock In {\em Proceedings of EMNLP\/}.

\bibitem[{Hashimoto et~al.(2017)Hashimoto, Xiong, Tsuruoka, and
  Socher}]{Hashimoto2017}
Kazuma Hashimoto, Caiming Xiong, Yoshimasa Tsuruoka, and Richard Socher. 2017.
\newblock {A Joint Many-Task Model: Growing a Neural Network for Multiple NLP
  Tasks}.
\newblock In {\em Proceedings of EMNLP\/}.

\bibitem[{Hinton et~al.(2015)Hinton, Vinyals, and Dean}]{Hinton2015}
Geoffrey Hinton, Oriol Vinyals, and Jeff Dean. 2015.
\newblock {Distilling the Knowledge in a Neural Network}.
\newblock {\em arXiv preprint arXiv:1503.02531\/} .

\bibitem[{Jacob et~al.(2009)Jacob, Vert, Bach, and Vert}]{Jacob2009}
Laurent Jacob, Jean-Philippe Vert, Francis~R Bach, and Jean-philippe Vert.
  2009.
\newblock {Clustered Multi-Task Learning: A Convex Formulation}.
\newblock In {\em Proceedings of NIPS\/}. pages 745--752.

\bibitem[{Jacobs et~al.(1991)Jacobs, Jordan, Nowlan, and Hinton}]{Jacobs1991}
Robert~a. Jacobs, Michael~I. Jordan, Steven~J. Nowlan, and Geoffrey~E. Hinton.
  1991.
\newblock {Adaptive Mixtures of Local Experts}.
\newblock {\em Neural Computation\/} 3(1):79--87.

\bibitem[{Kang et~al.(2011)Kang, Grauman, and Sha}]{Kang2011}
Zhuoliang Kang, Kristen Grauman, and Fei Sha. 2011.
\newblock {Learning with Whom to Share in Multi-task Feature Learning}.
\newblock In {\em Proceedings of ICML\/}.

\bibitem[{Kim et~al.(2015)Kim, Stratos, Sarikaya, and Jeong}]{Kim:ea:15}
Young-Bum Kim, Karl Stratos, Ruhi Sarikaya, and Minwoo Jeong. 2015.
\newblock {New Transfer Learning Techniques for Disparate Label Sets}.
\newblock In {\em Proceedings of ACL\/}.

\bibitem[{Kumar and {Daum{\'{e}} III}(2012)}]{Kumar2012}
Abhishek Kumar and Hal {Daum{\'{e}} III}. 2012.
\newblock {Learning Task Grouping and Overlap in Multi-task Learning}.
\newblock {\em Proceedings of the 29th International Conference on Machine
  Learning\/} pages 1383--1390.

\bibitem[{Kumar et~al.(2016)Kumar, Kohail, Kumar, Ekbal, and
  Biemann}]{Kumar2016}
Ayush Kumar, Sarah Kohail, Amit Kumar, Asif Ekbal, and Chris Biemann. 2016.
\newblock {IIT-TUDA at SemEval-2016 Task 5: Beyond Sentiment Lexicon: Combining
  Domain Dependency and Distributional Semantics Features for Aspect Based
  Sentiment Analysis}.
\newblock {\em Proceedings of SemEval\/} .

\bibitem[{Li and Zhou(2007)}]{Li:Zhou:07}
Ming Li and Zhi-Hua Zhou. 2007.
\newblock {Improve Computer-Aided Diagnosis With Machine Learning Techniques
  Using Undiagnosed Samples}.
\newblock {\em IEEE Transactions on Systems, Man and Cybernetics\/}
  37(6):1088--1098.

\bibitem[{Liu et~al.(2017)Liu, Qiu, and Huang}]{Liu2017}
Pengfei Liu, Xipeng Qiu, and Xuanjing Huang. 2017.
\newblock {Adversarial Multi-task Learning for Text Classification}.
\newblock In {\em Proceedings of ACL\/}.

\bibitem[{Luong et~al.(2016)Luong, Le, Sutskever, Vinyals, and
  Kaiser}]{Luong:ea:16}
Minh-Thang Luong, Quoc~V. Le, Ilya Sutskever, Oriol Vinyals, and Lukasz Kaiser.
  2016.
\newblock {Multi-task Sequence to Sequence Learning}.
\newblock In {\em Proceedings of ICLR\/}.

\bibitem[{Mohammad et~al.(2016)Mohammad, Kiritchenko, Sobhani, Zhu, and
  Cherry}]{mohammad-EtAl:2016:SemEval}
Saif Mohammad, Svetlana Kiritchenko, Parinaz Sobhani, Xiaodan Zhu, and Colin
  Cherry. 2016.
\newblock Semeval-2016 task 6: Detecting stance in tweets.
\newblock In {\em Proceedings of SemEval\/}.

\bibitem[{Nakov et~al.(2016)Nakov, Ritter, Rosenthal, Stoyanov, and
  Sebastiani}]{SemEval:2016:task4}
Preslav Nakov, Alan Ritter, Sara Rosenthal, Veselin Stoyanov, and Fabrizio
  Sebastiani. 2016.
\newblock {SemEval-2016 Task 4: Sentiment Analysis in Twitter}.
\newblock In {\em Proceedings of SemEval\/}. San Diego, California.

\bibitem[{Nangia et~al.(2017)Nangia, Williams, Lazaridou, and
  Bowman}]{Nangia2017}
Nikita Nangia, Adina Williams, Angeliki Lazaridou, and Samuel~R. Bowman. 2017.
\newblock {The RepEval 2017 Shared Task: Multi-Genre Natural Language Inference
  with Sentence Representations}.
\newblock In {\em Proceedings of RepEval\/}.

\bibitem[{Palogiannidi et~al.(2016)Palogiannidi, Kolovou, Christopoulou,
  Kokkinos, Iosif, Malandrakis, Papageorgiou, Narayanan, and
  Potamianos}]{palogiannidi2016tweester}
Elisavet Palogiannidi, Athanasia Kolovou, Fenia Christopoulou, Filippos
  Kokkinos, Elias Iosif, Nikolaos Malandrakis, Haris Papageorgiou, Shrikanth
  Narayanan, and Alexandros Potamianos. 2016.
\newblock {Tweester at SemEval-2016 Task 4: Sentiment Analysis in Twitter Using
  Semantic-Affective Model Adaptation}.
\newblock In {\em Proceedings of SemEval\/}. pages 155--163.

\bibitem[{Peng et~al.(2017)Peng, Thomson, Smith, and Allen}]{Peng2017}
Hao Peng, Sam Thomson, Noah~A Smith, and Paul~G Allen. 2017.
\newblock {Deep Multitask Learning for Semantic Dependency Parsing}.
\newblock In {\em Proceedings of ACL 2017\/}.

\bibitem[{Plank et~al.(2016)Plank, S{\o}gaard, and Goldberg}]{Plank2016a}
Barbara Plank, Anders S{\o}gaard, and Yoav Goldberg. 2016.
\newblock {Multilingual Part-of-Speech Tagging with Bidirectional Long
  Short-Term Memory Models and Auxiliary Loss}.
\newblock In {\em Proceedings of ACL\/}.

\bibitem[{Pontiki et~al.(2016)Pontiki, Galanis, Papageorgiou, Androutsopoulos,
  Manandhar, AL-Smadi, Al-Ayyoub, Zhao, Qin, De~Clercq, Hoste, Apidianaki,
  Tannier, Loukachevitch, Kotelnikov, Bel, Jim{\'e}nez-Zafra, and
  Eryi\u{g}it}]{Pontiki2016Aspect}
Maria Pontiki, Dimitris Galanis, Haris Papageorgiou, Ion Androutsopoulos,
  Suresh Manandhar, Mohammed AL-Smadi, Mahmoud Al-Ayyoub, Yanyan Zhao, Bing
  Qin, Orph{\'e}e De~Clercq, Veronique Hoste, Marianna Apidianaki, Xavier
  Tannier, Natalia Loukachevitch, Evgeniy Kotelnikov, N{\'u}ria Bel,
  Salud~Maria Jim{\'e}nez-Zafra, and G{\"u}l\c{s}en Eryi\u{g}it. 2016.
\newblock {SemEval-2016 Task 5: Aspect Based Sentiment Analysis}.
\newblock In {\em Proceedings of SemEval\/}.

\bibitem[{Rei(2017)}]{Rei2017}
Marek Rei. 2017.
\newblock {Semi-supervised Multitask Learning for Sequence Labeling}.
\newblock In {\em Proceedings of ACL 2017\/}.

\bibitem[{Riedel et~al.(2017)Riedel, Augenstein, Spithourakis, and
  Riedel}]{Riedel2017}
Benjamin Riedel, Isabelle Augenstein, Georgios~P Spithourakis, and Sebastian
  Riedel. 2017.
\newblock {A simple but tough-to-beat baseline for the Fake News Challenge
  stance detection task}.
\newblock In {\em arXiv preprint arXiv:1707.03264\/}.

\bibitem[{Riedel et~al.(2013)Riedel, Yao, McCallum, and Marlin}]{Riedel2013}
Sebastian Riedel, Limin Yao, Andrew McCallum, and Benjamin~M. Marlin. 2013.
\newblock {Relation Extraction with Matrix Factorization and Universal
  Schemas}.
\newblock {\em Proceedings of NAACL-HLT\/} pages 74--84.

\bibitem[{Ruder et~al.(2017)Ruder, Bingel, Augenstein, and
  S{\o}gaard}]{ruder2017sluice}
Sebastian Ruder, Joachim Bingel, Isabelle Augenstein, and Anders S{\o}gaard.
  2017.
\newblock {Sluice networks: Learning what to share between loosely related
  tasks}.
\newblock In {\em CoRR, abs/1705.08142\/}.

\bibitem[{Ruder et~al.(2016)Ruder, Ghaffari, and Breslin}]{Ruder2016a}
Sebastian Ruder, Parsa Ghaffari, and John~G. Breslin. 2016.
\newblock {A Hierarchical Model of Reviews for Aspect-based Sentiment
  Analysis}.
\newblock {\em Proceedings of EMNLP\/} pages 999--1005.

\bibitem[{Ruder and Plank(2017)}]{ruder2017emnlp}
Sebastian Ruder and Barbara Plank. 2017.
\newblock {Learning to select data for transfer learning with Bayesian
  Optimization}.
\newblock In {\em Proceedings of EMNLP\/}.

\bibitem[{S\o{}gaard and Goldberg(2016)}]{Soegaard:Goldberg:16}
Anders S\o{}gaard and Yoav Goldberg. 2016.
\newblock Deep multi-task learning with low level tasks supervised at lower
  layers.
\newblock In {\em Proceedings of ACL\/}.

\bibitem[{Vo and Zhang(2015)}]{Vo2015}
Duy-Tin Vo and Yue Zhang. 2015.
\newblock {Target-Dependent Twitter Sentiment Classification with Rich
  Automatic Features}.
\newblock In {\em Proceedings of IJCAI\/}. pages 1347--1353.

\bibitem[{Xue et~al.(2007)Xue, Liao, Carin, and Krishnapuram}]{Xue2007}
Ya~Xue, Xuejun Liao, Lawrence Carin, and Balaji Krishnapuram. 2007.
\newblock {Multi-Task Learning for Classification with Dirichlet Process
  Priors}.
\newblock {\em Journal of Machine Learning Research\/} 8:35--63.

\bibitem[{Yeh et~al.(2017)Yeh, Wu, Ko, and Wang}]{Yeh:ea:17}
Chih-Kuan Yeh, Wei-Chieh Wu, Wei-Jen Ko, and Yu-Chiang~Frank Wang. 2017.
\newblock {Learning Deep Latent Space for Multi-Label Classification}.
\newblock In {\em Proceedings of AAAI\/}.

\bibitem[{Yu et~al.(2005)Yu, Tresp, and Schwaighofer}]{Yu2005}
Kai Yu, Volker Tresp, and Anton Schwaighofer. 2005.
\newblock {Learning Gaussian processes from multiple tasks}.
\newblock {\em Proceedings of ICML\/} 22:1012--1019.

\bibitem[{Zhang et~al.(2012)Zhang, Reichart, Barzilay, and
  Globerson}]{Zhang:ea:12}
Yuan Zhang, Roi Reichart, Regina Barzilay, and Amir Globerson. 2012.
\newblock {Learning to Map into a Universal POS Tagset}.
\newblock In {\em Proceedings of EMNLP\/}.

\end{thebibliography}
\bibliographystyle{acl_natbib}

\end{document}